\documentclass[letterpaper]{article} 
\usepackage{aaai23}  
\usepackage{times}  
\usepackage{helvet}  
\usepackage{courier}  
\usepackage[hyphens]{url}  
\usepackage{graphicx} 
\usepackage{natbib}  
\usepackage{caption} 
\usepackage{algorithm}
\usepackage{algorithmic}

\usepackage{amsmath}
\usepackage{booktabs}
\usepackage{multirow}
\usepackage[normalem]{ulem}
\usepackage{amssymb}
\usepackage{newfloat}
\usepackage{listings}
\DeclareCaptionStyle{ruled}{labelfont=normalfont,labelsep=colon,strut=off} 
\floatstyle{ruled}
\newfloat{listing}{tb}{lst}{}
\floatname{listing}{Listing}
\title{Improving Scene Text Image Super-resolution via \\ Dual Prior Modulation Network}

\author{
    Shipeng Zhu\textsuperscript{\rm 1,2}, Zuoyan Zhao\textsuperscript{\rm 1,2}, Pengfei Fang\textsuperscript{\rm 1,2}, Hui Xue\textsuperscript{\rm 1,2}\thanks{Corresponding author.}
}

\affiliations{
    \textsuperscript{\rm 1}School of Computer Science and Engineering, Southeast University, Nanjing 210096, China\\
    \textsuperscript{\rm 2}MOE Key Laboratory of Computer Network and Information Integration (Southeast University), China\\
    
    \{shipengzhu, zuoyanzhao, fangpengfei, hxue\}@seu.edu.cn
}

\begin{document}

\maketitle

\begin{abstract}
Scene text image super-resolution (STISR) aims to simultaneously increase the resolution and legibility of the text images, and the resulting images will significantly affect the performance of downstream tasks. Although numerous progress has been made, existing approaches raise two crucial issues: (1) They neglect the global structure of the text, which bounds the semantic determinism of the scene text. (2) The priors, e.g., text prior or stroke prior, employed in existing works, are extracted from pre-trained text recognizers. That said, such priors suffer from the domain gap including low resolution and blurriness caused by poor imaging conditions, leading to incorrect guidance. Our work addresses these gaps and proposes a plug-and-play module dubbed Dual Prior Modulation Network (DPMN), which leverages dual image-level priors to bring performance gain over existing approaches. Specifically, two types of prior-guided refinement modules, each using the text mask or graphic recognition result of the low-quality SR image from the preceding layer, are designed to improve the structural clarity and semantic accuracy of the text, respectively. The following attention mechanism hence modulates two quality-enhanced images to attain a superior SR result. Extensive experiments validate that our method improves the image quality and boosts the performance of downstream tasks over five typical approaches on the benchmark. Substantial visualizations and ablation studies demonstrate the advantages of the proposed DPMN. Code is available at: https://github.com/jdfxzzy/DPMN.
\end{abstract}

\section{Introduction}
\label{sec:intro}

Scene text images, containing rich linguistic and graphic information, are widely present in our daily life. The understanding of scene text images is an integral part of various high-level applications, like scene text recognition~\cite{fang2021read}, scene text retrieval~\cite{wang2021scene}, and text-based image captioning~\cite{zhang2022magic}. However, limited zone in the image and inadequate imaging conditions~\cite{long2021scene} cause an issue of low resolution (LR) for texts. This, in turn, leads to unreliable text understanding in such images, thereby degrading the performance of downstream tasks. In this context, it is necessary to develop a super-resolution (SR) method, to recover LR scene text images to high-quality SR ones.

\begin{figure}[t]
\centering
\includegraphics[width=0.95\columnwidth]{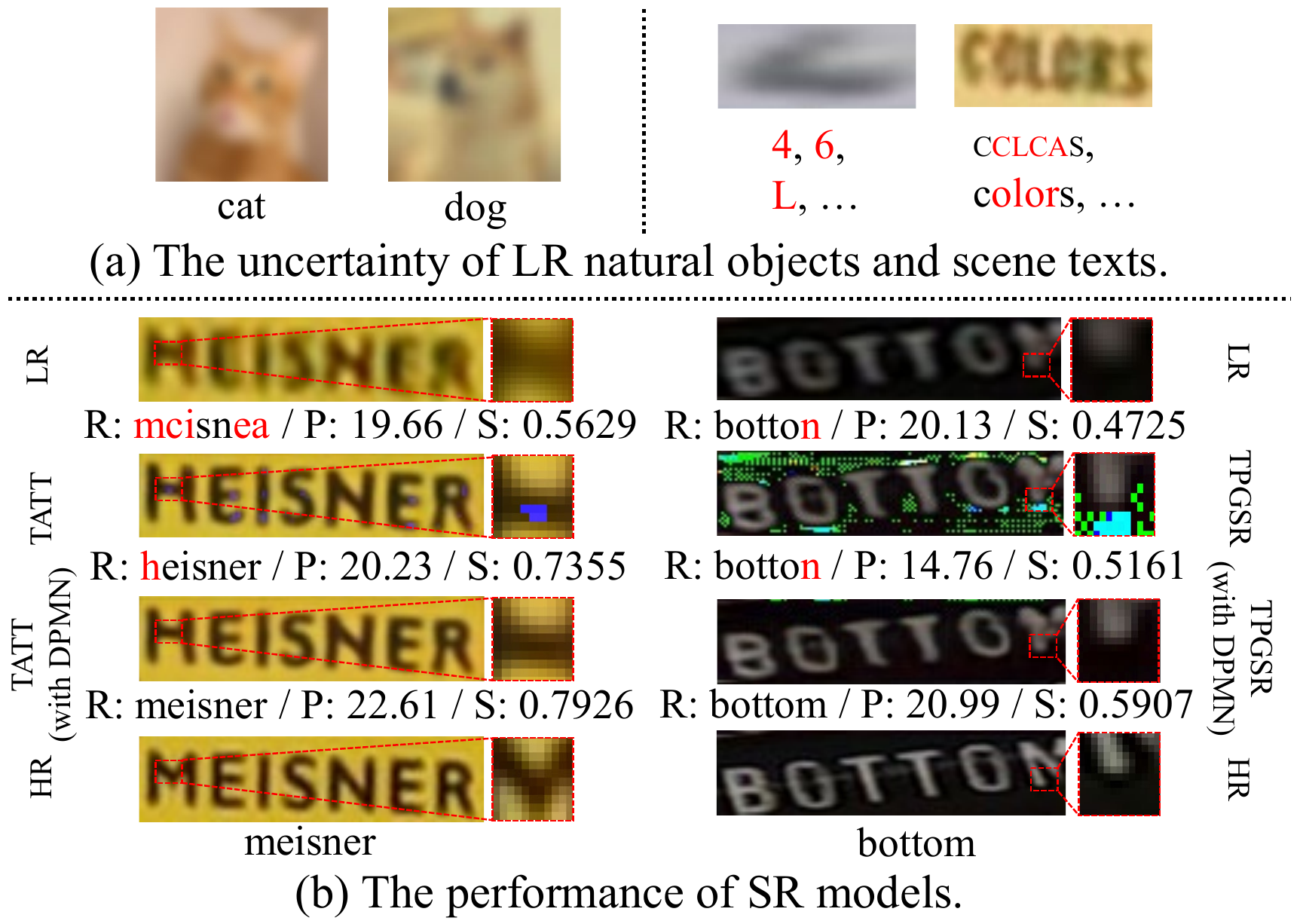} 
\caption{The illustration of effect about global structure and graphic semantic priors on the scene text image. ``R", ``P" and ``S" represent recognition results, PSNR, and SSIM.}
\label{fig:title}
\end{figure}

To achieve this goal, some early attempts~\cite{xu2017learning,pandey2018binary} simply utilize techniques of conventional SR methods, e.g., $L_{1}$ loss, to improve the quality of scene text in images. However, they cannot effectively boost the performance of downstream tasks. Subsequently, some methods, tailored for the scene text image super-resolution (STISR) task, benefit from the superficial properties of the scene text. For example, the pioneering work, Text Super-Resolution Network (TSRN), tends to perceive the sequential information of the text via CNN-BiLSTM layers~\cite{wang2020scene}. The most recent works attempt to leverage various text properties from LR images, as prior, to steer the SR process. Text Gestalt (TG) model~\cite{chen2022text} uses the local stroke structure to capture the stroke-aware prior from a Transformer-based recognizer to prompt the training phase. Text Prior Guided Super-Resolution (TPGSR) model~\cite{ma2021text} and the following Text ATTention network (TATT)~\cite{ma2022text} leverage the pre-trained text recognizer to obtain a text prior, i.e., the probability sequence of a scene text image whose length denotes the number of characters learned by the text recognizer. This hence improves the quality of SR images as well as the performance of downstream tasks.

Despite the significant advances that have been made, existing methods ignore two essential facts, limiting their further improvement. \textbf{First, the global structure information of scene text plays a vital role in STISR}. In terms of the scene text, global structure contains character strokes and the orientation of the text sequence (see HR images in Figure~\ref{fig:title}(b)). Unlike natural objects, scene text in the image is the spatial cluster of discrete characters, where each item is a continuous graphic~\cite{ye2014text}. Moreover, the semantic definiteness of the text is determined by the global structure containing characteristics at multi scales~\cite{yao2014strokelets}. That said, natural objects like animals can be recognized by local structure and texture, even omitting most of the global structure information, e.g., shape and profile. However, missing such a global structure brings uncertainty to the scene text (see Figure~\ref{fig:title}(a))~\cite{wu2019editing}. As a direct representation of the global structure, the text mask has been ignored or solely used as the fourth channel of the input for enhancement by existing methods~\cite{wang2020scene, ma2022text}. \textbf{Second, the recognizer-based priors have inherent shortcomings}. Concretely, current prior-guided methods merely employ a pre-trained text recognizer to extract the text prior, such that the domain gap of resolution will result in incorrect prior information. Meanwhile, we empirically observe that such methods are prone to generate artifacts. As shown in Figure~\ref{fig:title}(b), TATT and TPGSR produce amiss images, and characters are falsely predicted by the recognizer, which again shows the necessity of the structure prior to the scene text image.

\begin{figure}[t]
\centering
\includegraphics[width=1\columnwidth]{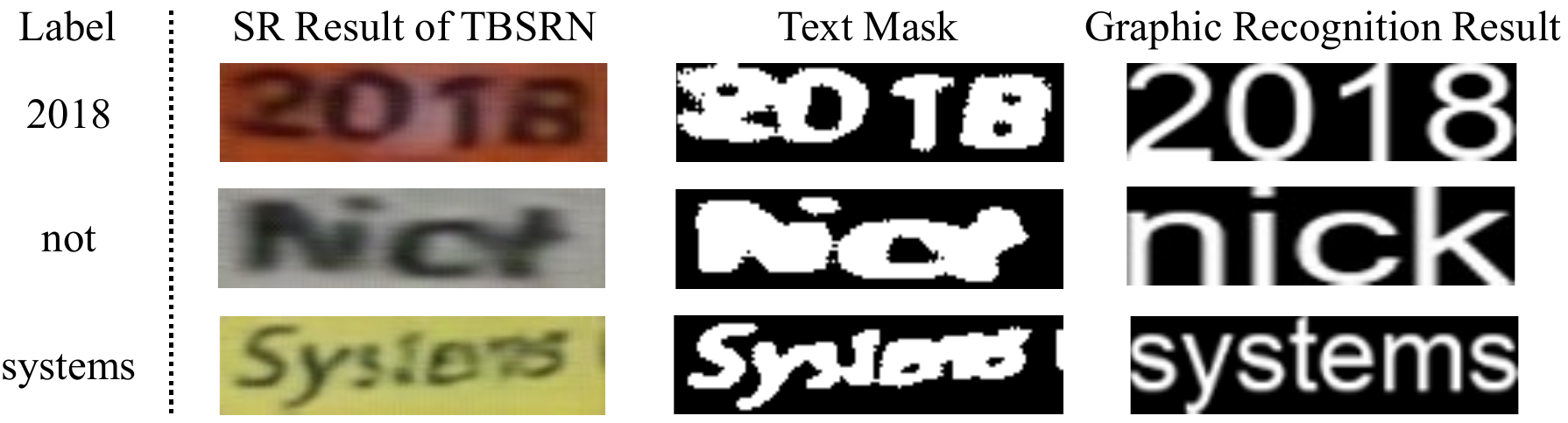} 
\caption{The illustration of global structure and graphic semantic priors in an existing STISR approach. }
\label{fig:title2}
\end{figure}

To address the issues above, we utilize two explicit image-level priors, i.e., text mask and graphical recognition result from low-quality images. These two priors provide complementary information. Specifically, the text mask brings superiority in terms of the global structure, while the graphical recognition result contributes to clear semantic features. That is, the former compensates for the deficiency of character correctness and attributes w.r.t the graphical property, while the latter compensates for the ambiguity of the mask in terms of local essential information. This induction is shown in Figure~\ref{fig:title2}.
Having those two priors in mind, we propose Dual Prior Modulation Network (DPMN), a plug-and-play module that enjoys the global structural information and the local semantic information to improve the quality of the SR images produced by existing models. In doing so, two branches of Prior-Guided Refinement Modules (PGRM) are designed to create the text mask prior and the graphical recognition prior, and each processes SR images guided by the global structural information and the local semantic information, respectively. A following Complementation Modulation Module (CMM) is further proposed to modulate and fuse the reconstructed SR images, refined by two PGRM branches. Of note, our work can be understood as a post-processor of STISR networks, such that it can be seamlessly used in existing established works. Our \textbf{contributions} are summarized as follows: 

\begin{figure*}[t]
\centering
\includegraphics[width=1\textwidth]{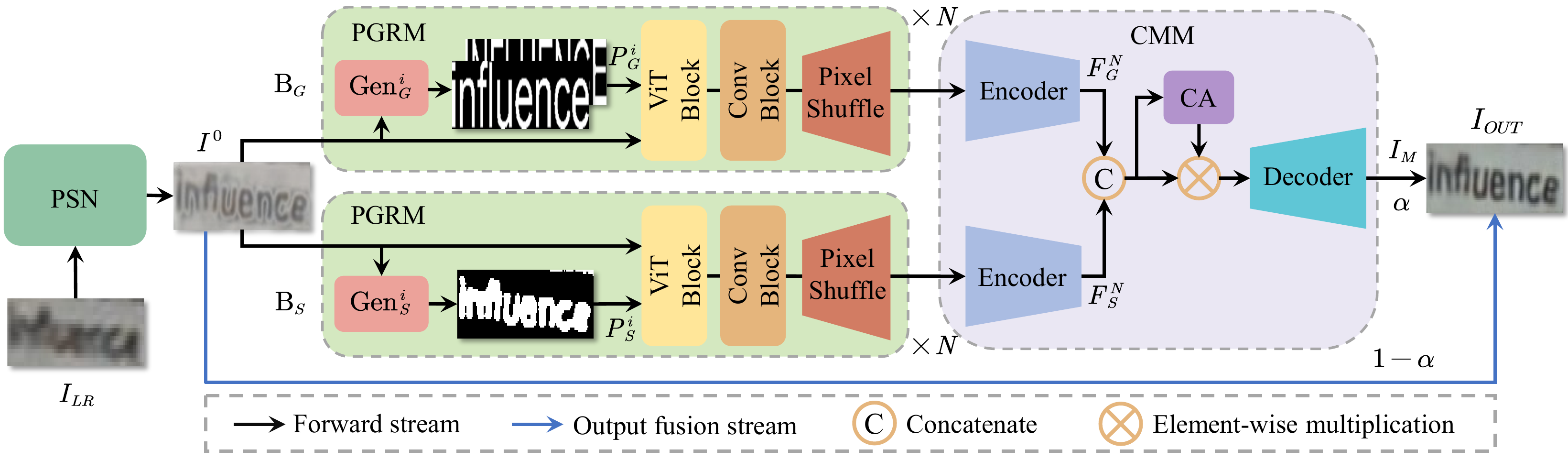} 
\caption{The overall architecture of our proposed Dual Prior Modulation Network (DPMN). It consists of two main modules:  Prior-Guided Refinement Modules (PGRMs) and Complementation Modulation Module (CMM). ``Gen", ``Conv" and ``CA" denote Prior Generator, Convolutional Block, and Channel-wise Attention Block, respectively. The input of our DPMN is provided by the Pre-trained STISR Network (PSN) with frozen parameters.}
\label{fig:archi}
\end{figure*}

\begin{itemize}
\item We propose a novel Dual Prior Modulation Network (DPMN), which leverages the text mask and the graphical recognition result as priors, to improve the quality of SR images. By doing so, DPMN benefits from the global structure and semantic information, attaining a superior SR result.

\item In DPMN, each PGRM generates a prior and produces the improved SR image via interacting the scene text image and the prior image. Then enhanced SR images are modulated and fused by a CMM.

\item Extensive experiments demonstrate that the proposed DPMN can boost image quality and the performance of the text recognition task on the TextZoom benchmark on top of existing methods. Additional analysis exhibits the generalizability of DPMN.

\end{itemize}

\section{Related Work}

\subsection{Single Image Super-Resolution}

Single image super-resolution (SISR) aims to recover HR images from LR ones~\cite{wang2020deep}. The pioneering work, called SRCNN~\cite{dong2015image}, employs CNNs to regress a complex non-linear mapping to reconstruct the HR images and achieves superior performance. This is the first attempt in the deep learning area. Subsequent approaches improve the quality of SR outputs by developing advanced learning strategies and neural architectures. To enrich the knowledge of the SR model, the content-based approaches develop transfer learning algorithms, delivering knowledge from pre-trained classification networks to SR networks~\cite{ledig2017photo, johnson2016perceptual}.
As another way to learn informative features, a growing number of approaches develop the attention mechanism~\cite{fang2021attention} to attend to useful regions on LR images~\cite{zhang2018image, dai2019second, mei2021image}. Recent studies also show the Transformer with a self-attention mechanism gains promising performance in SISR~\cite{chen2021pre,liang2021swinir}.

\subsection{Scene Text Image Super-Resolution}

Unlike the SISR task, scene text image super-resolution (STISR) is a more challenging task, which imposes requirements for understanding the text in images. The initial solutions utilize prior statistical knowledge to guide the SR process. In~\cite{capel2000super}, the maximum a posterior (MAP) method is adopted to predict new pixels in SR images. The Teager Filter~\cite{mancas2005super} employs the Taylor decomposition to highlight high frequencies of the text.
Recently, many works exploit the properties of the scene text to improve SR networks. For example, TSRN~\cite{wang2020scene} and PCAN~\cite{zhao2021scene} apply the CNN-BiLSTM module to perceive sequential features of the scene text. The prior information is also considered as essential auxiliary information for the SR process. Specifically, TBSRN~\cite{chen2021scene} benefits from the supervision of character-level features, which are developed by a pre-trained text recognizer. The TG model~\cite{chen2022text} uses the local structure prior, i.e., text strokes, to improve the quality of SR images. TPGSR~\cite{ma2021text} and the following TATT~\cite{ma2022text} further show that priors with text semantic information are also beneficial for the STISR task.

Although significant progress has been made, existing methods suffer from the lack of constraints on global structure information and imprecise priors from the recognizer, which may limit further performance improvements. Inspired by this, our work proposes a dual-branch network to produce superior SR images, benefiting from complementary priors.

\section{Methodology}

This section first provides a sketch of the proposed Dual Prior Modulation Network (DPMN). Then we continue to present a detailed description of two units in DPMN, i.e., the Prior-Guided Refinement Module (PGRM) and the Complementation Modulation Module (CMM). We will also introduce the training objective of the proposed network.


\subsection{Overall Architecture}

The overall architecture of the proposed DPMN is illustrated in Figure~\ref{fig:archi}. Our proposed DPMN is built on top of existing STISR networks. For any pre-trained STISR network, denoted by $\mathrm{PSN}$ in Figure~\ref{fig:archi}, it first receives LR images $I_{{LR}} \in \mathbb{R}^{h \times w \times 3}$ as input, and produces primary SR images $I^{0} = \mathrm{PSN}(I_{{LR}}) \in \mathbb{R}^{2h \times 2w \times 3}$. Then the DPMN further refines $I^{0}$ using two branches of networks, with each consisting of $N$ Prior-Guided Refinement Modules (PGRMs). Notably, the two branches, called $\mathrm{B}_{G}$ and $\mathrm{B}_{S}$, refine $I^{0}$ guided by the graphic semantic prior $P_G$ and the global structure prior $P_S$,
respectively. For the $i$-th PGRM in branch $\mathrm{B}_{G}$, a generator $\mathrm{Gen}^i_{G}$ first generates a graphic semantic prior $P^{i}_{G}$, given the refined images $I_{G}^{i-1}$ from the previous PGRM as input. Then both $I_{G}^{i-1}$ and $P^{i}_{G}$ are sent to the following refinement modules to produce a newly refined image $I_{G}^{i}$. In the same vein, the PGRM in branch $\mathrm{B}_{S}$ has a similar workflow using the global structure prior $P_{S}$. Having the refined images $I_{G}^{N}$ and $I_{S}^{N}$ in hand, the Complementation Modulation Module (CMM) aggregates the two images and produces a modulated one $I_{M}$. Of note, the parameters of the PSN are frozen while only the DPMN is optimized in the training phase. In the inference phase, we employ a fusion strategy to ensure the robustness of the final output, i.e., $I_{OUT}= \alpha \times I_{M} + (1-\alpha) \times I^{0}$~ for $0< \alpha <1$.

\subsection{Prior-Guided Refinement Module}

Two critical issues arise for the refinement process guided by priors in the STISR task: (1) How to mitigate the influence of imprecise prior information due to defective input (see Figure~\ref{fig:title2}). (2) How to integrate useful prior information, as guidance, into images. Our Prior-Guided Refinement Module (PGRM) is proposed to address those two issues. This is achieved by calibrating the low-quality SR input by information interaction of the input and the prior. Each PGRM (see Figure~\ref{fig:archi}) contains a prior generator, a Vision Transformer (ViT) block, a convolutional block, and a pixel shuffle layer.

As shown in Figure~\ref{fig:title2}, we can empirically find that the global structure of the scene text and graphic semantics can provide complementary guidance for the restoration of the SR image. This motivates us to exploit two representative image-level information as priors, i.e., the text mask and the graphic recognition result.

To utilize these two priors, we propose to extract and use the priors in a parallel manner, such that each branch of PGRM processes a specific prior to the fullest. Different from the feature map or the embedding vector~\cite{ma2021text,chen2021scene}, the prior in graphic-recognition-guided branch $\mathrm{B}_{G}$ (see Figure~\ref{fig:archi}) can present the semantic information of text images. This can avoid bringing the noise, due to the domain gap, from the ambiguous semantics to the input images. For the $i$-th PGRM in $\mathrm{B}_{G}$, a pre-trained recognizer first produces a text result for the input $I^{i-1}_{G}$, and the following rendering module~\cite{gupta2016synthetic} further transfers the text result to an image-format data containing upper and lower case letters, as the graphic prior, denoted by $P^{i}_{G}$. This processing is formulated as:
\begin{equation}
P^{i}_{G}=\mathrm{Ren}(\mathrm{Rec}(I^{i-1}_{G})) \in \mathbb{R}^{2h \times 2w \times 2}.
\label{eq:pg}
\end{equation}

In the $i$-th PGRM unit of the global-structure-guided branch $\mathrm{B}_{S}$, we utilize the text mask generated by the binarization operation of $I^{i-1}_{S}$ as the structure prior $P^{i}_{S}$. The structure information, e.g., font, size, tendency, et al, can be represented by $P^{i}_{S}$. Its operation is given by:
\begin{equation}
P^{i}_{S}=\mathrm{Bin}(I^{i}_{S}) \in \mathbb{R}^{2h \times 2w \times 1}.
\label{eq:ps}
\end{equation}



\begin{figure}[t]
\centering
\includegraphics[width=1\columnwidth]{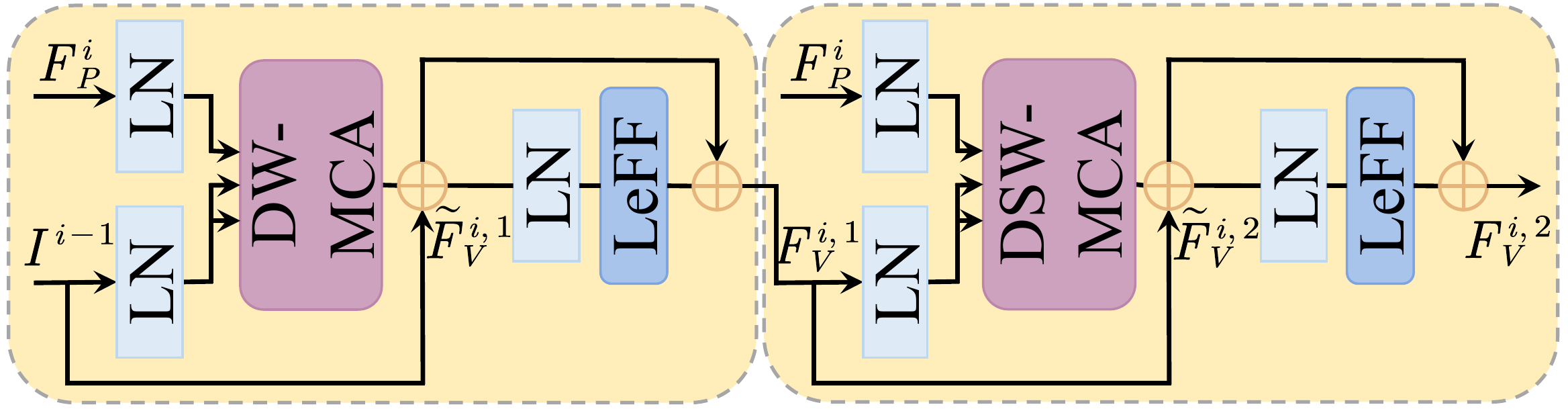}

\caption{The architecture of the Vision Transformer (ViT) block in $i$-th PGRM of each branch. ``DW-MCA" and ``DSW-MCA" are short for Dynamic Window Multi-head Cross Attention and Dynamic Shifted Window Multi-head Cross Attention. ``LN" and ``LeFF" denote Layer Normalization and Locally-enhanced Feed-Forward Network.}
\label{fig:ViTBlock}
\end{figure}

Existing STISR approaches simply stack the Sequential Residual Block (SRB) in the backbone network, to capture and mine the sequence dependence of the text. However, the SRB has difficulty to understand the deformable text due to the rigid nature of the Bi-LSTM. Meanwhile, most prior-based approaches manually fuse the prior information into the input image via simple operations, e.g., addition. In our work, we want the network to learn the fusion protocol adaptively, thereby improving the quality of the SR images.

We address this by interacting the information flow between the prior and the input image via the ViT. Before formulating the information interaction, we first perform the dimensionality matching operation. That is, since the dimensions of the input image are located in the space of $\mathbb{R}^{2h \times 2w \times 3}$, we need to project the two priors (e.g., $P^{i}_{G}$ and $P^{i}_{S}$) to the same space. For $P^{i}_{G} \in \mathbb{R}^{2h \times 2w \times 2}$, we use a convolutional layer to do the transformation, obtaining input feature $F^i_{P_{G}}=\mathrm{Conv}(P^i_{G}) \in \mathbb{R}^{2h \times 2w \times 3}$. For the global structure prior $P^{i}_{S}$, we expand it in the channel wise, as $F^i_{P_{S}} \in \mathbb{R}^{2h \times 2w \times 3}$.

As shown in Figure~\ref{fig:ViTBlock}, the ViT block in the $i$-th PGRM of arbitrary branch is realized by the cross attention mechanism~\cite{zhou2020cross}, which takes the prior $F^{i}_{P}$ (collective term for $F^{i}_{P_G}$ and $F^{i}_{P_S}$) as the query, and inputs low-quality SR image $I^{i-1}$ (collective term for $I^{i-1}_{G}$ and $I^{i-1}_{S}$) as the key and value, to perform the information interaction. Following the recent DW-ViT~\cite{ren2022beyond}, we also process the ViT block in two stages. In the first stage, a multi-head cross attention layer (denoted by DW-MCA) with residual connection receives the $F^i_{P}$ and $I^{i-1}$ as input, to produce multi-scale correlation by three sizes of windows, given by:
\begin{equation}
\tilde{F}_{V}^{i,1}=\mathrm{DW\mbox{-}MCA}(\mathrm{LN}(F_{P}^{i}), \mathrm{LN}(I^{i-1}))+\mathrm{LN}(I^{i-1}).
\label{eq:DW-MCA}
\end{equation}

Along with interacting the prior and the image with dynamic windows mechanism, we also enable the DW-ViT the capacity to capture local contextual information for $F_{V}^{i,1}$, realized by:
\begin{equation}
F_{V}^{i,1}=\tilde{F}_{V}^{i,1}+\mathrm{LeFF}\left( \mathrm{LN}\left( \tilde{F}_{V}^{i,1} \right) \right), 
\label{eq:FSR}
\end{equation}
where $\mathrm{LeFF}$ is the locally-enhanced feed-forward network with spatial-wise and depth-wise convolutional layer proposed in UFormer~\cite{wang2022uformer}.

In the second stage, we replace the DW-MCA layer with the dynamic shifted window multi-head cross attention (DSW-MCA) layer to build long-range interaction. This stage can be formulated as:
\begin{equation}
\tilde{F}_{V}^{i,2}=\mathrm{DSW\mbox{-}MCA}(\mathrm{LN}(F_{P}^{i}), \mathrm{LN}(F_{V}^{i,1}))+\mathrm{LN}(F_{V}^{i,1}),
\label{eq:Hi2}
\end{equation}
\begin{equation}
F_{V}^{i,2}=\tilde{F}_{V}^{i,2}+\mathrm{LeFF}\left( \mathrm{LN}\left( \tilde{F}_{V}^{i,2} \right) \right).
\label{eq:FSRi1}
\end{equation}

Recent researches have proved that the ViT has its internal drawbacks, e.g., the limitation of encoding the inductive bias, for vision tasks~\cite{li2022contextual}. Thereby, a convolutional block (denoted by $\mathrm{Conv}$), which consists of two convolutional layers, is added after the ViT block. This design can improve the perception ability of locality and spatial invariance~\cite{peng2021conformer} in PGRM. Subsequently, the widely-used pixel shuffle layer ($\mathrm{PS}$)~\cite{shi2016real} produces the refined image, as:
\begin{equation}
I^{i}= \mathrm{PS}(\mathrm{Conv}(F_{V}^{i,2})).
\label{eq:IiSR}
\end{equation}

To optimize the proposed PGRM block efficiently, we adopt two image-level losses. The pixel loss $\mathcal{L}_{pix}$ constrains the information of image content, and the gradient profile loss $\mathcal{L}_{gp}$ considers to minimize the information loss of edge details~\cite{wang2020scene}. For the $i$-th PGRM module, the loss function is given by:
\begin{equation}
\mathcal{L} _{img}^{i}=\lambda _p\underset{\mathcal{L} _{pix}}{\underbrace{||I_{HR}-I^{i}||_2}}+\lambda _g\underset{\mathcal{L} _{gp}}{\underbrace{||\nabla I_{HR}-\nabla I^{i}||_1}}.
\label{eq:Limg}
\end{equation}

Since each branch includes $N$ PGRM blocks, the total loss of each branch is:
\begin{equation}
\mathcal{L}_{B} = \sum_{i=1}^{N}{\mathcal{L}^{i}_{img}}.
\label{eq:branch}
\end{equation}

\subsection{Complementation Modulation Module}

After the enhancement to the quality of SR images in two branches, we have two refined SR images, i.e., $I^{N}_{{G}}$ and $I^{N}_{{S}}$. A neural module is required to fuse and modulate those two images to attain a superior SR one. As shown in Figure~\ref{fig:archi}, this is realized by an encoder-decoder architecture. Each encoder is composed of six convolutional layers to extract the key features, as $F^{N}_{G}=\mathrm{Encoder}(I^{N}_{G})$ and $F^{N}_{S}=\mathrm{Encoder}(I^{N}_{S})$. Then we concatenate two feature maps, as:
\begin{equation}
F^{N}_{M} = \mathrm{Concat}(F^{N}_{G},F^{N}_{S}).
\label{eq:FN}
\end{equation}

We then employ the channel attention mechanism~\cite{hu2018squeeze} to learn the modulation importance per slice in $F^{N}_{M}$ and weight $F^{N}_{M}$ in channel-wise. A symmetric decoder is further used to produce the final modulated SR images. This is summarized as:
\begin{equation}
I_{M} = \mathrm{Decoder}(\mathrm{CA}(F^{N}_{M}) \otimes F^{N}_{M} + F^{N}_{M}).
\label{eq:ISR}
\end{equation}

In CMM, we use $\mathcal{L}_{img}$ to restrain $I_{M}$ by HR images $I_{HR}$, which can be described as follows:
\begin{equation}
\mathcal{L}_{CMM} = ||I_{HR}-I_{M}||_{2} + ||\nabla I_{HR}-\nabla I_{M}||_{1}.
\label{eq:fusion}
\end{equation}

\subsection{Training Objective}
In the training phase, we optimize the parameters of the proposed DPMN. The objective loss includes two types of components, i.e., the branch loss $\mathcal{L}_B$ and the CMM loss $\mathcal{L}_{CMM}$. The total loss function is:
\begin{equation}
\mathcal{L}_{Total} =  \lambda_{C}\mathcal{L}_{CMM} + \lambda_{G} \mathcal{L}_{B_G}+ \lambda_{S} \mathcal{L}_{B_S}.
\label{eq:total}
\end{equation}

\begin{table*}[t]\small
\centering
\renewcommand\arraystretch{1}
\begin{tabular}{c|cccc|cccc|cccc}
\hline
\multirow{2}{*}{Method} & \multicolumn{4}{c|}{ASTER} & \multicolumn{4}{c|}{CRNN} & \multicolumn{4}{c}{MORAN} \\ \cline{2-13} 
 & Easy & Medium & Hard & Average & Easy & Medium & Hard & Average & Easy & Medium & Hard & Average \\ \hline
TSRN & 73.32 & 56.20 & 39.17 & 57.31 & 54.73 & 41.25 & 32.24 & 43.47 & 67.88 & 49.96 & 37.08 & 52.64 \\
+DPMN & \textbf{74.43} & \textbf{56.41} & \textbf{39.24} & \textbf{57.81} &\textbf{54.91} & \textbf{41.46} & \textbf{32.46} & \textbf{43.68} & \textbf{68.07} & \textbf{50.18} & \textbf{37.16} & \textbf{52.80} \\ \hline
TBSRN & 76.71 & 59.53 & 43.71 & 61.03 & 59.79 & 45.07 & 34.18 & 47.18 & 70.17 & 55.63 & 40.80 & 56.46 \\
+DPMN & \textbf{76.78} & \textbf{60.52} & \textbf{44.08} & \textbf{61.49} & 59.79 & \textbf{45.22} & \textbf{34.40} & \textbf{47.29} & \textbf{70.54} & 55.63 & \textbf{40.95} & \textbf{56.64} \\ \hline
TG & 77.02 & 62.58 & 42.37 & 61.72 & 59.42 & 48.62 & 34.33 & 48.23 & 72.58 & 57.97 & 39.76 & 57.79 \\
+DPMN & \textbf{77.39} & \textbf{62.72} & \textbf{42.96} & \textbf{62.08} & \textbf{59.48} & \textbf{48.76} & \textbf{34.40} & \textbf{48.32} & \textbf{72.88} & \textbf{58.19} & \textbf{40.21} & \textbf{58.11} \\ \hline
TPGSR & 78.01 & 60.67 & 42.67 & 61.56 & 58.62 & 45.92 & 33.88 & 46.92 & 72.33 & 55.71 & 39.91 & 57.01 \\
+DPMN & \textbf{78.13} & 60.67 & \textbf{42.74} & \textbf{61.63} & \textbf{59.36} & 45.92 & \textbf{34.03} & \textbf{47.24} & \textbf{72.95} & \textbf{56.13} & \textbf{40.06} & \textbf{57.42} \\ \hline
TATT & 78.51 & 63.29 & 44.97 & 63.30 & 64.30 & 54.15 & 39.09 & 53.28 & 72.88 & 61.02 & 43.78 & 60.12 \\
+DPMN & \textbf{79.25} & \textbf{64.07} & \textbf{45.20} & \textbf{63.89} & \textbf{64.36} & 54.15 & \textbf{39.24} &  \textbf{53.35} & \textbf{73.26} & \textbf{61.45} & \textbf{43.86} &  \textbf{60.42} \\ \hline
HR & 93.39 & 86.96 & 75.65 & 85.87 & 76.41 & 75.05 & 64.56 & 72.33 & 89.01 & 83.13 & 71.11 & 81.62 \\ \hline
\end{tabular}
\caption{The recognition accuracy ($\%$) on TextZoom. The bold numbers denote the better score between the baseline and improved method by DPMN.}
  \label{tab:RecognitionResults}
\end{table*}

\section{Experiments}

In this section, we first introduce the experiment datasets, evaluation metrics, and implementation details. Then we conduct comparison experiments and ablation studies to demonstrate the superiority of our method. 

\subsection{Evaluation Datasets and Metrics}

The STISR benchmark \textbf{TextZoom}~\cite{wang2020scene} is collected in real-world scenarios. It consists of 17,367 LR-HR image pairs for training and 4,373 pairs for testing. Wherein, the test set is divided into three subsets to indicate different levels of blurriness, i.e., easy (1,619 pairs), medium (1,411 pairs) and hard (1,343 pairs). The size of LR images is $16 \times 64$, while the size of HR images is $32 \times 128$.

We use the peak signal-to-noise ratio (PSNR) and the structural similarity index measure (SSIM) metrics to evaluate the quality of the SR images. In order to measure the downstream task performance, we calculate the recognition accuracy for the text recognition task. Following the common practice in~\cite{wang2020scene}, the recognition results are evaluated on ASTER~\cite{shi2018aster}, CRNN~\cite{shi2016end}, MORAN~\cite{luo2019moran} models.

\subsection{Baselines and Implementation Details}
We evaluate the proposed DPMN on five STISR models as baselines, including TSRN~\cite{wang2020scene}, TBSRN~\cite{chen2021scene}, TPGSR~\cite{ma2021text}, TG~\cite{chen2022text}, and TATT~\cite{ma2022text}. 
In our experiments, we directly use models from the official implementation or perform the identical hyper-parameters as reported in the official implementations to train the baseline models.
Of note, to understand the net improvement from the proposed DPMN, we respectively select three models with the best performance on ASTER, CRNN, and MORAN, as baselines of each aforementioned STISR method. Then we fix the parameters of the SR models and only train the DPMNs. 

We implement our model with PyTorch 1.10 deep learning library~\cite{paszke2019pytorch} and all the experiments are conducted on one RTX 3090 GPU. For each experiment, the DPMN is trained 20 epochs using Adam optimizer~\cite{kingma2014adam}. The learning rate is set to 0.001, and the size of the mini-batch is 48. We empirically observe that the loss function is insensitive to the parameter $\lambda$, and we set all the $\lambda$ to 1. In the inference phase, the output fusion ratio $\alpha$ is selected based on the pre-trained baselines. The size of original SR results and modulated images is $32 \times 128$. We apply the pre-trained VisionLANs~\cite{wang2021two} in PGRMs as the text prior generators. In terms of the network architecture, the number of PGRMs in each branch, $N$, is set to 3. In the ViT block, the window numbers of the DW-MCA and the DSW-MCA are 2, 4, and 8 with patch size 2, while the head number of the MCA is set to 6. Additionally, we exploit the adaptive dense-connection~\cite{xie2019adaptive} and the self-distillation mechanism~\cite{zhang2019your} to ensure the stability of training and speed up the convergence rate.

\subsection{Experimental Results}

\begin{table}[t]
\centering
\begin{tabular}{cccc}
\hline
Method & PSNR & SSIM & Accuracy \\ \hline
TSRN & 20.81 & 0.7594 & 51.14 \\
+DPMN & \textbf{21.09} & \textbf{0.7698} & \textbf{51.43} \\ \hline
TBSRN & 20.91 & 0.7625 & 54.89 \\
+DPMN & \textbf{21.11} & \textbf{0.7650} & \textbf{55.14} \\ \hline
TG & 18.80 & 0.6597 & 55.91 \\
+DPMN & \textbf{20.56} & \textbf{0.7472} & \textbf{56.17} \\ \hline
TPGSR & 21.18 & 0.7615 & 55.16 \\
+DPMN & \textbf{21.33} & \textbf{0.7718} & \textbf{55.43} \\ \hline
TATT & 21.21 & 0.7825 & 58.90 \\
+DPMN & \textbf{21.49} & \textbf{0.7925} & \textbf{59.22} \\ \hline
\end{tabular}
\caption{The average image quality scores and average recognition accuracy (\%) on TextZoom.}
  \label{tab:Imagequality}
\end{table}

\begin{figure*}[ht]
\centering
\includegraphics[width=1\textwidth]{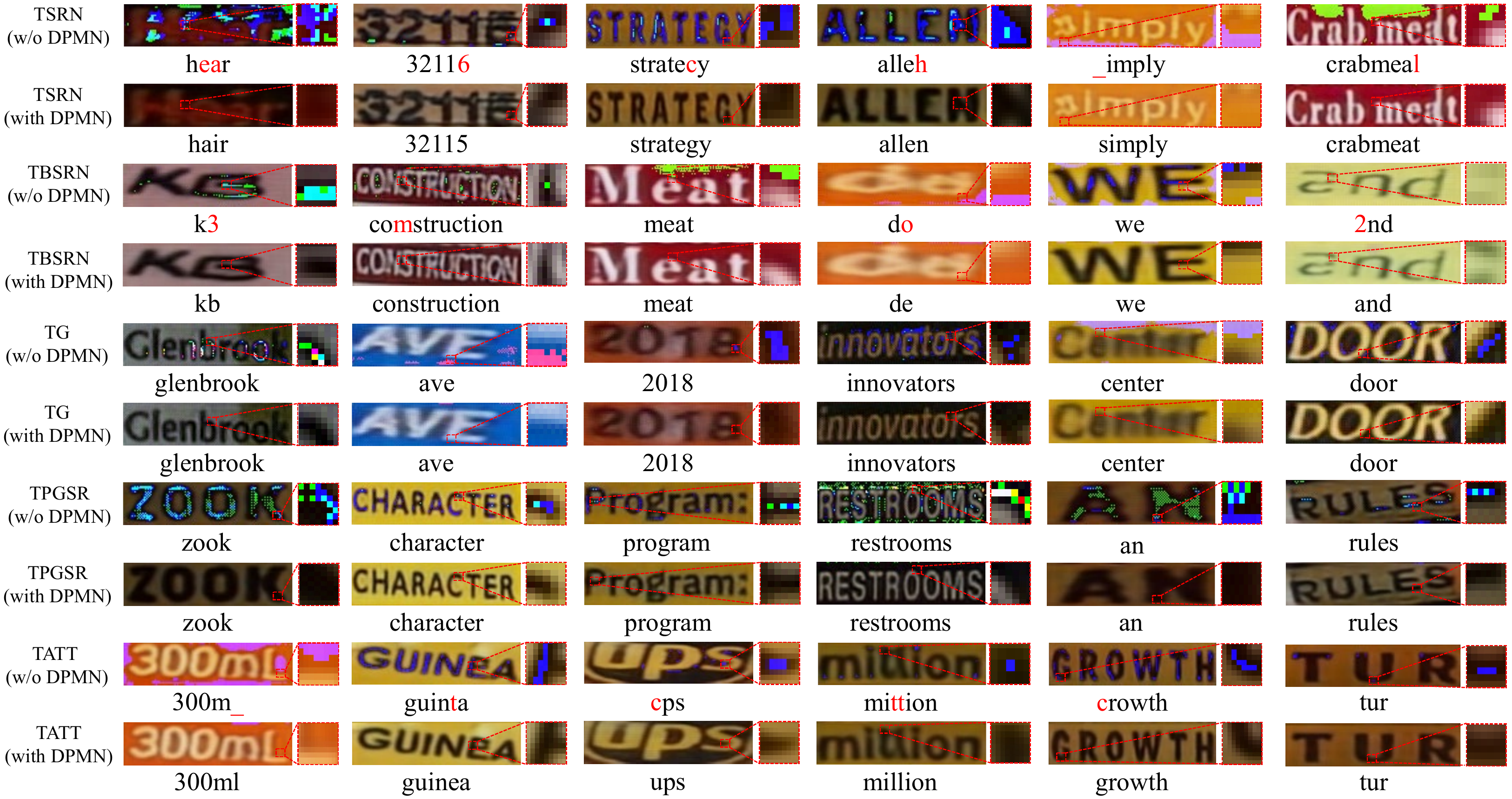}
\caption{The SR images with recognition results on TextZoom based on ASTER. The red characters mean wrong results.}
\label{fig:vis}
\end{figure*}

We conduct quantitative experiments on the benchmark TextZoom. The comparison results are shown in Table~\ref{tab:RecognitionResults} and Table~\ref{tab:Imagequality}. We can clearly observe that our method boosts the image quality and the recognition accuracy of existing baselines even if the data is derived from the best models obtained on different text recognizers. Taking the latest TATT as an example, our method achieves better recognition accuracy and image quality simultaneously (improves average Accuracy/PSNR/SSIM by 0.32/0.28/0.01). 
Compelled by the observations above, our proposed DPMN has the universal enhancement capability for existing STISR models given the all-around improvement across all metrics.

Meanwhile, we provide some qualitative studies by comparing the visualization of the SR images between the baselines and our work. The results and comparisons shown in Figure~\ref{fig:vis} reveal that: (1) Existing methods have defects w.r.t. color reproduction and structure retention. In contrast, our method can produce precise details, verifying the potential of the complementary prior information. (2) The baseline models are prone to generate artifacts, which degrade the image quality and the performance of the downstream recognition task, while DPMN can avoid the artifacts due to explicit guidance from the text structure information.

\subsection{Ablation Study}

In this section, we conduct ablation studies to investigate the effectiveness of motivation and model components. All the evaluations are validated on TextZoom. The default baseline is TATT based on ASTER.

\begin{table}[t]
\centering
\resizebox{1\linewidth}{!}{
\begin{tabular}{c|c|c|c|c}
\hline
Setting & Prior & PSNR & SSIM & Accuracy \\ \hline
\multirow{3}{*}{SingleB} & Mask & 21.12 & 0.7864 & 63.50 \\
 & GRR & 19.59 & 0.7472 & 63.62 \\
 & Mask+GRR & 21.01 & 0.7859 & 63.67 \\ \hline
\multirow{2}{*}{DualB} & Mask\&GRR & \textbf{21.49} & \textbf{0.7925} & \textbf{63.89} \\
 & Mask\&GRR$^{\ast}$ & 23.60 & 0.8875 & 77.57 \\ \hline
\end{tabular}}
\caption{The performance of different priors. ``SingleB" denotes Single-Branch, ``DualB" denotes Dual-Branch, ``GRR" denotes Graphic Recognition Results, ``+" denotes concatenate, and ``$^{\ast}$" denotes the dual priors are from the HR image.}
\label{tab:Branch}
\end{table}

\subsubsection{Effect of the Two Priors}
We first perform experiments to demonstrate the necessity of complementary priors in the SR process. This study has two settings, i.e., single-branch and dual-branch. The results in Table~\ref{tab:Branch} convince us that: (1) The mask prior performs better in image quality, while the graphic recognition result, including rich semantic features, shows superior performance in the text recognition accuracy. This justifies our motivation for modulating the two complementary priors. (2) The dual-branch setting outperforms the single-branch one, clearly showing our design is reasonable and superior. Specifically, we obtain the two priors from HR images, which are impossible to be acquired in the natural SR process, and evaluate the upper bound of the image quality enhancement. The stunning results again indicate that the two priors play a promising role in modulation.

\begin{table}[t]
\centering
\resizebox{1\linewidth}{!}{
\begin{tabular}{c|cc|cccc}
\hline
 & PSN & DPMN & Easy & Medium & Hard & Average \\ \hline
(\romannumeral1) & Frozen & None & 78.51 & 63.29 & 44.97 & 63.30 \\
(\romannumeral2) & None & Train & 66.09 & 45.92 & 32.99 & 49.42 \\
(\romannumeral3) & Fine-tune & Train & 79.18 & 63.08 & 44.38 & 63.30 \\ \hline
(\romannumeral4) & Frozen & Train & \textbf{79.25} & \textbf{64.07} & \textbf{45.20} & \textbf{63.89} \\ \hline
\end{tabular}}
\caption{The recognition accuracy (\%) of different training strategies. ``PSN" represents the Pre-trained STSIR Network. ``None" denotes not using this module. ``Fine-tune" denotes fine-turning based on a pre-trained model. ``Frozen" denotes freezing the parameters of the pre-trained model.}
  \label{tab:PlugandPlay}
\end{table}

\subsubsection{Effect of the Training Strategy}
In this study, we investigate the training strategy of the proposed DPMN. Table~\ref{tab:PlugandPlay} shows the results of possible training strategies. We can find that training the DPMN, on top of the pre-trained PSN with fixed parameters, achieves the best performance, showing the flexibility and effectiveness of DPMN. Notably, applying DPMN independently (see (\romannumeral2)) as an image SR backbone leads to trivial results. One conjecture may lie in that the ambiguous priors from LR images struggle to provide proper guidance for the SR process, thereby leading to problematic initialization and weak complementary information.

\begin{table}[t]
\centering
\begin{tabular}{c|cccc}
\hline
$N$ & Easy & Medium & Hard & Average \\ \hline
1 & 78.81 & \textbf{64.07} & 45.12 & 63.71 \\
2 & 78.69 & 63.78 & 45.12 & 63.57 \\
4 & 78.57 & 64.00 & 45.12 & 63.60 \\
5 & 78.63 & 63.86 & \textbf{45.20} & 63.60 \\ \hline
3 & \textbf{79.25} & \textbf{64.07} & \textbf{45.20} & \textbf{63.89} \\ \hline
\end{tabular}
\caption{The recognition accuracy ($\%$) of different PGRM numbers in each branch.}
  \label{tab:NumberofPGRM}
\end{table}

\begin{table}[t]
\centering
\begin{tabular}{ccl|cccc}
\hline
\multicolumn{3}{c|}{Variant} & Easy & Medium & Hard & Average \\ \hline
\multicolumn{1}{c|}{\multirow{3}{*}{FW}} & \multicolumn{2}{c|}{2} & 78.69 & 63.78 & \textbf{45.20} & 63.59 \\
\multicolumn{1}{c|}{} & \multicolumn{2}{c|}{4} & 78.81 & 63.57 & 45.12 & 63.55 \\
\multicolumn{1}{c|}{} & \multicolumn{2}{c|}{8} & 79.06 & 63.64 & 45.05 & 63.64 \\ \hline
\multicolumn{3}{c|}{DW (DPMN)} & \textbf{79.25} & \textbf{64.07} & \textbf{45.20} & \textbf{63.89} \\ \hline
\end{tabular}
\caption{The recognition accuracy ($\%$) of different multi-head cross attention layers. ``FW" denotes Fixed Window, whose size is set to the subsequent number. ``DW" denotes Dynamic Window.}
  \label{tab:DesignofWindow}
\end{table}

\begin{table}[t]
\centering
\begin{tabular}{c|cccc}
\hline
CMM & Easy & Medium & Hard & Average \\ \hline
Encoder-decoder & 78.88 & 63.43 & \textbf{45.20} & 63.55 \\
U-Net & 78.94 & 63.29 & 45.05 & 63.48 \\
TSRN & 78.75 & 63.08 & 45.05 & 63.34 \\ \hline
DPMN & \textbf{79.25} & \textbf{64.07} & \textbf{45.20} & \textbf{63.89} \\ \hline
\end{tabular}
\caption{The recognition accuracy ($\%$) comparison between different variants of CMM.}
  \label{tab:DesignofCMM}
\end{table}

\subsubsection{Number of PGRMs}
We implement the proposed DPMN in a symmetric fashion, e.g., each branch containing three PGRMs. In this study, we evaluate the effect of the number of PGRMs. The results are reported in Table~\ref{tab:NumberofPGRM}. We can empirically find that when $N=3$, the proposed DPMN achieves the overall best performance over three settings, as well as the peak value of the average accuracy. We use $N=3$ as the default value for the experiments in this paper.

\subsubsection{Effect of the Dynamic Window Mechanism}
In this part, we study the dynamic window mechanism in the proposed DPMN. The results in Table~\ref{tab:DesignofWindow} show that: (1) The large window helps achieve better performance in the easy subset, while the MCA with a small window performs admirably in the hard subset. (2) Our method exhibits remarkable results in all subsets, which demonstrates the effectiveness of the dynamic window mechanism. These observations verify that the dynamic window can understand the images with different blur levels, thereby enriching its learning capacity.

\subsubsection{Design of the CMM}
We utilize the encoder-decoder architecture with a channel-wise attention (CA) module to fuse the two refined SR images. In this study, we empirically compare it with three methods, i.e., single encoder-decoder without CA mechanism, U-Net~\cite{ronneberger2015u} with skip connection, and TSRN~\cite{wang2020scene} from previous STISR. Table~\ref{tab:DesignofCMM} also shows that the architecture of CMM in our work outperforms other vanilla variants, again showing the superiority of our design.

\section{Conclusion}
In this paper, we propose a Dual Prior Modulation Network (DPMN) to boost the performance of existing Scene Text Image Super-Resolution (STISR) methods. We leverage the global text structure and graphical semantics as complementary priors to guide the SR image refinement progressively in dual branches. This is realized by the Prior-Guided Refinement Module (PGRM) and the Complementation Modulation Module (CMM). Substantial experiments and ablation studies demonstrate the effectiveness of DPMN, which improves both image quantity and the performance of the downstream recognition task. 
We believe our work will provide valuable intuition for further improvement of the STISR task. Future work will focus on developing a more efficient and effective backbone network for the STISR task.

\section{Acknowledgments}
This work was supported by National Natural Science Foundation of China (Grant No.62076062) and Collaborative Innovation Center of Wireless Communications Technology.

\bibliography{aaai23}

\end{document}